\documentclass{article}
\usepackage{spconf,amsmath,graphicx,hyperref}
\usepackage{xspace}
\usepackage{booktabs}
\usepackage[table]{xcolor}

\newcommand{\ourmethod}{{\fontfamily{lmtt}\selectfont \textbf{TTARAG}}\xspace}
\DeclareMathOperator*{\argmax}{arg\,max}

\title{Predict the Retrieval! Test Time Adaptation for Retrieval Augmented Generation}
\name{Xin Sun$^{1}$ Zhongqi Chen$^{2}$ Qiang Liu$^{1}$ Shu Wu$^{1}$ Bowen Song$^{2}$ Weiqiang Wang$^{2}$ Zilei Wang$^{3}$ Liang Wang$^{1}$}
\address{$^{1}$ NLPR, MAIS, CASIA $^{2}$ Ant Group $^{3}$ USTC\\ xin.sun@cripic.ia.ac.cn,  \{qiang.liu, shu.wu, wangliang\}@nlpr.ia.ac.cn, zlwang@ustc.edu.cn, \\
 \{chenzhongqi.czq, bowen.sbw, weiqiang.wwq\}@antgroup.com}
\begin{document}
\ninept
\maketitle
\begin{abstract}
Retrieval-Augmented Generation (RAG) has emerged as a powerful approach for enhancing large language models' question-answering capabilities through the integration of external knowledge. However, when adapting RAG systems to specialized domains, challenges arise from distribution shifts, resulting in suboptimal generalization performance. In this work, we propose \ourmethod, a test-time adaptation method that dynamically updates the language model's parameters during inference to improve RAG system performance in specialized domains. Our method introduces a simple yet effective approach where the model learns to predict retrieved content, enabling automatic parameter adjustment to the target domain. Through extensive experiments across six specialized domains, we demonstrate that \ourmethod achieves substantial performance improvements over baseline RAG systems. Code available at \url{https://github.com/sunxin000/TTARAG}.
\end{abstract}
\section{Introduction}
Retrieval-Augmented Generation (RAG) \cite{izacard2021leveraging, lewis2020retrieval, edge2024local, sun2025dividethenalign} has emerged as a crucial approach for enhancing large language models (LLMs) \cite{radford2019language, brown2020language, sun2025kbqar1} by addressing their inherent knowledge limitations. Through the integration of external knowledge sources \cite{jin2019pubmedqa}, RAG systems not only improve the accuracy of LLM responses but also help mitigate hallucination issues while eliminating the need for extensive model retraining.

However, while most current research has focused on the effectiveness of RAG systems for general domains, significant challenges persist in adapting these systems to specialized domains. These systems often struggle with distribution shifts and domain-specific data dependencies \cite{xu2025simragselfimprovingretrievalaugmentedgeneration, shi2024medadapter, sun2024dive}, frequently failing to accurately utilize information in domain-specific contexts \cite{miller2020effect, liu-etal-2022-challenges}. 

To address these challenges, test-time adaptation (TTA) \cite{sun2020test} offers a promising solution. TTA allows models to dynamically adjust their parameters at inference time, typically through self-supervised learning objectives derived from the test data itself, without requiring new labeled data \cite{liang2024comprehensive, sun2023noiserobust}. This characteristic makes TTA particularly well-suited for scenarios involving domain shifts or distribution changes not seen during initial training, as it enables the model to rapidly specialize to the immediate context. Building on these insights, we propose a simple yet powerful method for adapting RAG systems during inference: \ourmethod. Our approach generates self-supervised learning signals by dividing retrieved passages into prefix-suffix pairs and training the model to predict suffix content from prefix context. This technique enables LLMs to perform real-time parameter updates when encountering new domains, effectively leveraging domain knowledge stored within the model parameters.

Through extensive experiments across six specialized domains, we demonstrate that \ourmethod achieves substantial performance improvements over baseline RAG systems. Our approach consistently outperforms both standard RAG and  baselines like Chain-of-Thought and In-Context Learning, achieving the best results in 19 out of 24 experimental settings while maintaining computational efficiency. These results validate the effectiveness of our approach for domain-specific applications.

\section{Related Work}

\subsection{Retrieval Augmented Generation}
Retrieval Augmented Generation (RAG) \cite{lewis2020retrieval} has emerged as a powerful paradigm for enhancing large language models (LLMs) with external knowledge. By integrating a retrieval system with LLMs, RAG enables models to access and leverage external knowledge sources during generation, effectively addressing the limitations of static, parameterized knowledge in LLMs.

Recent advances in RAG have focused on several key directions. First, researchers have explored dynamic retrieval processes \cite{jiang-etal-2023-active,jeong-etal-2024-adaptive} to improve the relevance of retrieved content. Second, various filtering mechanisms \cite{robustlm,yu2024rankrag} have been developed to eliminate irrelevant contexts and enhance RAG robustness. Additionally, instruction-tuning methods \cite{liu2024chatqa,lin2024radit} have been specifically designed to improve LLMs' search and RAG capabilities. Recent work has also explored knowledge boundary alignment \cite{sun2025dividethenalign}, uncertainty-aware long-context modeling \cite{li2024uncertaintyrag}, and reinforcement learning for question answering \cite{sun2025kbqar1,sun2026gapd}.
\subsection{Test-time inference adaptation}

Test-time inference adaptation aims to adapt pre-trained models to unlabeled test data during inference time without accessing the source training data. This paradigm has gained increasing attention as a practical solution for handling distribution shifts in real-world applications \cite{wang2021tent, sun2024dive}. Unlike traditional domain adaptation methods that require simultaneous access to both source and target domains, test-time adaptation only needs the pre-trained model and target data, making it more privacy-friendly and storage-efficient \cite{liang2020we}.

Early works in this direction focused on hypothesis transfer learning, where models trained on source domains are adapted to target domains with limited labeled data. Recent advances have extended this to fully unsupervised scenarios, leveraging techniques like entropy minimization \cite{wang2021tent} and self-training \cite{sun2020test} to adapt models using only unlabeled test samples. 

Building on these advances, \ourmethod introduces a simple yet effective approach for test-time adaptation in retrieval-augmented generation. By learning to predict subsequent tokens in retrieved passages, \textbf{\ourmethod enables fully unsupervised adaptation without requiring access to source domain data or labeled examples.}

\section{Methodology}
Our approach introduces a test-time adaptation mechanism for retrieval-augmented generation that enables model optimization during inference without access to ground truth labels. The key innovation lies in designing a self-supervised learning objective using retrieved passages as supervision signals. We hypothesize that by training the model to predict a suffix of a passage given its prefix and the original query, the model is encouraged to better understand the contextual relationships and domain-specific language patterns within the retrieved documents. This task inherently pushes the model to learn how information flows and connects within relevant texts, thereby aligning its internal representations more closely with the nuances of the target domain encountered at test time.

\subsection{Overview}
Given a test input query $q$ and retrieved passages $\{p_1, ..., p_k\}$, we formulate a self-supervised adaptation objective by splitting passages into prefix-suffix pairs for prediction:
\begin{equation}
    \mathcal{L}_{adapt} = -\sum_{i=1}^k \log P(p_i^{suffix}|p_i^{prefix}, q; \theta)
\end{equation}
where $\theta$ represents the model parameters.

\begin{figure}[h]
    \centering
    \includegraphics[width=1.0\linewidth]{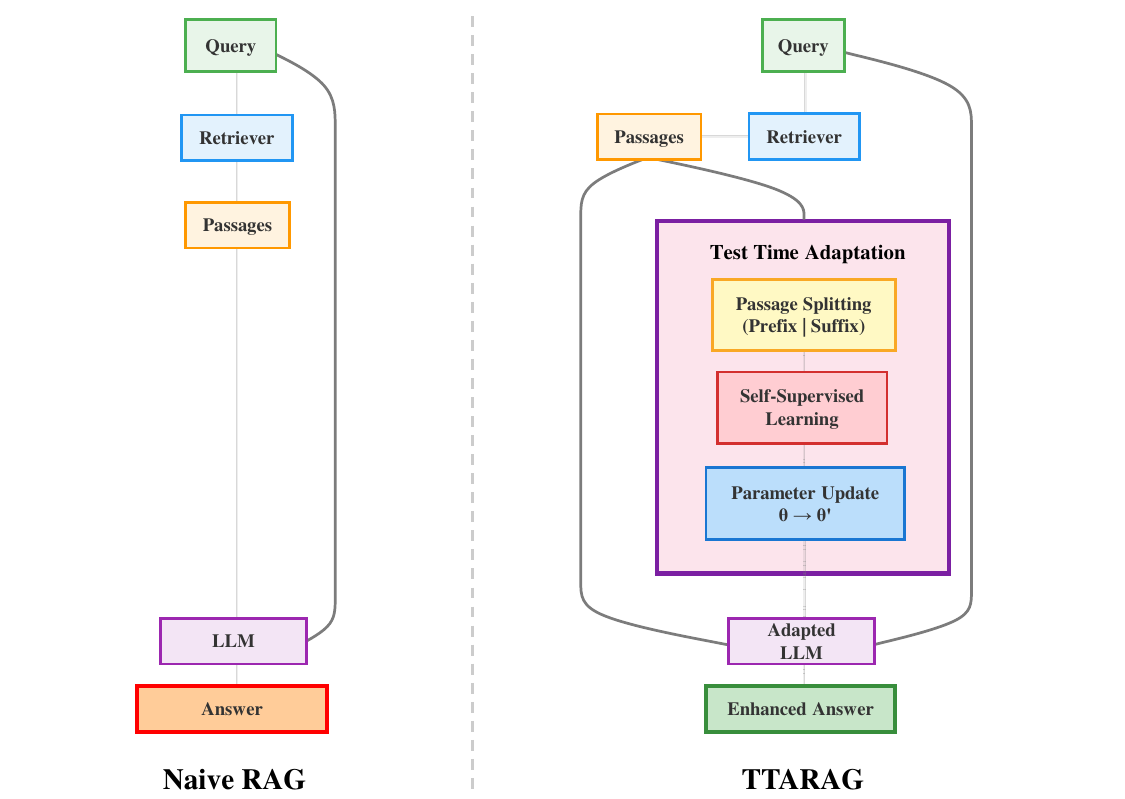}
    \caption{Comparison between Standard RAG and TTARAG systems.}
    \label{fig:overview}
\end{figure}

\subsection{Context Processing}
The adaptation process begins with careful processing of the retrieved passages to create meaningful prefix-suffix pairs for training. 

\noindent\textbf{Length Filtering.} To ensure sufficient context for meaningful adaptation, passages shorter than a configured minimum length threshold are filtered out.

\noindent\textbf{Passage Splitting.} Each passage is split into prefix-suffix pairs using a two-tier strategy:
    \begin{itemize}
        \item \textbf{Primary Strategy} Passages are split at first natural linguistic boundaries marked by punctuation (periods, commas, semicolons, colons, exclamation marks, and question marks)
        \item \textbf{Fallback Strategy} When no suitable punctuation-based split exists, the passage is divided at its midpoint, ensuring each segment contains at least three words.
    \end{itemize}

\subsection{Parameter Adaptation Process}

\subsubsection{Initialization}
Prior to the adaptation process, the model parameters are reset to their original pre-trained state to ensure a clean starting point for each adaptation iteration. 


\subsubsection{Training Loop}
For each batch of prefix-suffix pairs:
\begin{equation}
    \theta_t = \theta_{t-1} - \alpha \cdot \frac{1}{N} \sum_{i=1}^N \nabla_\theta \mathcal{L}_{adapt}^i
\end{equation}
where $N$ is the gradient accumulation steps and $\mathcal{L}_{adapt}^i$ is the loss for the $i$-th pair.

During training, the complete text (prefix and suffix) is first tokenized. The model then computes the loss on the suffix prediction task, where prefix tokens are masked during label preparation. To ensure stable training, gradients are accumulated over two steps and clipped to a maximum norm threshold. The AdamW optimizer then updates model parameters using these accumulated gradients. Since we only adapt on 3 prefix-suffix pairs in our experiments, the computational overhead remains acceptable.

\subsection{Response Generation}
After parameter adaptation, the model generates the final response using the adapted parameters $\theta'$:
\begin{equation}
    y = \argmax_y P(y|q, \{p_1, ..., p_k\}; \theta')
\end{equation}


\section{Experiments}

\begin{table*}[t]
  \footnotesize
    \centering
    \caption{Performance comparison across different domains. Numbers represent accuracy scores (\%). Best results for each model group are shown in \textbf{bold}. naive-rag refers to the respective LLM with standard retrieval-augmentation (top 20 passages) but without \ourmethod adaptation.}
    \label{tab:results}
    \begin{tabular}{@{}lcccccccc@{}}
    \toprule
    & \multicolumn{6}{c}{\textbf{CRAG}} & \multicolumn{2}{c}{\textbf{Medical}} \\
    \cmidrule(lr){2-7} \cmidrule(l){8-9}
    \textbf{Model} & \textbf{Finance} & \textbf{Sports} & \textbf{Music} & \textbf{Movie} & \textbf{Open} & \textbf{Overall} & \textbf{BioASQ} & \textbf{PubMed} \\
    \midrule
    \multicolumn{9}{l}{\textit{Llama-3.1-8b-it}} \\
    \cmidrule(l){1-9}
    naive-rag & 17.4 & 27.6 & 34.9 & 31.3 & 42.4 & 29.8 & 55.6 & 46.6 \\
    +CoT & 17.9 & \textbf{30.2} &37.6 & 31.5 & \textbf{45.8} & 31.6 & 54.6 & 50.8 \\
    +ICL & 16.1 & 24.8 & 33.5 & 29.4 & 40.4 & 28.0 & 49.8 & 53.6 \\
    \rowcolor{gray!20}\ourmethod & \textbf{20.1} & 29.5 & \textbf{37.7} & \textbf{34.6} & 41.5 & \textbf{31.9} & \textbf{75.0} &  \textbf{57.4}  \\
    \rowcolor{gray!20}$\Delta$ vs naive-rag & +2.7 & +1.9 & +2.8 & +3.3 & -0.9 & +2.1 & +19.4 & +10.8 \\
    \midrule
    
    \multicolumn{9}{l}{\textit{Llama-2-7b-chat}} \\
    \cmidrule(l){1-9}
    naive-rag & 14.7 & 23.2 & 36.5 & 30.4 & 39.2 & 27.8 & 54.1 & 47.6 \\
    +CoT & 15.7 & \textbf{26.7} & 34.3 & 31.4 & \textbf{41.5} & 29.1 & 55.1 & 48.2 \\
    +ICL & 16.0 & 24.2 & 36.1 & 31.2 & 39.2 & 28.4 & 55.6 & 43.4 \\
    \rowcolor{gray!20}\ourmethod & \textbf{16.4} & 25.8 & \textbf{40.7} & \textbf{33.8} & 41.1 & \textbf{30.5} & \textbf{71.8} & \textbf{54.0} \\
    \rowcolor{gray!20}$\Delta$ vs naive-rag & +1.7 & +2.6 & +4.2 & +3.4 & +1.9 & +2.7 & +17.7 & +6.4 \\
    \midrule
    \multicolumn{9}{l}{\textit{ChatGLM-3-6b}} \\
    \cmidrule(l){1-9}
    naive-rag & 9.8 & 18.7 & 31.4 & 22.4 & 33.4 & 22.0 & 51.4 & 19.8 \\
    +CoT & 12.7 & 20.6 & 28.4 & \textbf{25.8} & 33.9 & 23.6 & 44.3 & 22.4 \\
    +ICL & 9.9 & 18.2 & 30.8 & 22.1 & 33.0 & 21.8 & 50.8 & 19.2 \\
    \rowcolor{gray!20}\ourmethod & \textbf{14.0} & \textbf{22.1} & \textbf{33.5} & 25.5 & \textbf{38.1} & \textbf{25.7} & \textbf{58.4} & \textbf{44.8} \\
    \rowcolor{gray!20}$\Delta$ vs naive-rag & +4.2 & +3.4 & +2.1 & +3.1 & +4.7 & +3.7 & +7.0 & +25.0 \\
    \bottomrule
    \end{tabular}
    \end{table*}

    \begin{table}[h]
        \centering
        \caption{Number of samples in each domain of CRAG dataset.}
        \label{tab:dataset_stats}
        \begin{tabular}{lccccc}
        \toprule
        \textbf{Domain} & Finance & Sports & Music & Movie & Open \\
        \midrule
        \#Samples & 661 & 519 & 373 & 611 & 542 \\
        \bottomrule
        \end{tabular}
        \end{table}

\subsection{Datasets}
We conduct experiments on CRAG \cite{yang2024crag} as the evaluation benchmark. CRAG is a comprehensive RAG benchmark containing 2,706 question-answer pairs across five domains: Finance, Sports, Music, Movie, and Open domain. The questions are constructed through web content-based creation where annotators formulate real-world questions answerable through web search. For retrieval, CRAG provides each question with up to 50 
full HTML pages retrieved through the Brave Search API, with 
the top 5 pages showing a retrieval recall of approximately 
69\%.  Then MiniLM-v6 is employed to reranker all the sentence in the HTML and select top 20 sentences as the retrieval content. The statistics of the CRAG dataset are shown in Table~\ref{tab:dataset_stats}.

To evaluate the effectiveness of \ourmethod in the medical domain, we conduct additional experiments on two specialized datasets: PubMedQA \cite{jin2019pubmedqa}, which contains 1,000 biomedical research question-answer pairs, and BioASQ \cite{bioasq}, comprising 500 expert-curated question-answer pairs from the biomedical literature. For these medical datasets, as they do not come with pre-retrieved passages, we employed a standard BM25 retriever implemented using the Pyserini library to fetch the top 5 relevant passages from their respective corpora (PubMed abstracts for PubMedQA and biomedical articles for BioASQ). This retriever setup was also consistently applied when evaluating baseline models on these datasets.

\subsection{Evaluation Metrics}
 Instead of relying on exact match (EM)—which is sensitive to paraphrasing, aliases, and formatting variations—we adopt the CRAG evaluation protocol and employ an LLM-as-a-judge (Qwen2.5-72B-Instruct) to determine semantic equivalence between model predictions and reference answers. This LLM-based judging approach emphasizes the underlying meaning rather than surface-level string similarity, enabling robust assessment of whether the model output conveys the same information as the gold answer. In our evaluation, the LLM-judger demonstrates 98\% agreement with human annotators on a random sample of 100 examples, indicating high reliability and consistency. Overall, this methodology provides a more accurate and meaningful evaluation of model performance compared to traditional exact match metrics.

\subsection{Baselines}
We evaluate \ourmethod against several strong baselines, including prompting techniques (\textbf{Chain-of-Thought} \cite{wei2022chain}, \textbf{In-Context Learning} \cite{brown2020language}) and state-of-the-art pretrained RAG models (\textbf{Ret-Robust} 
 \cite{robustlm}, \textbf{RAAT} \cite{fang2024enhancing} , \textbf{Self-RAG} \cite{asai2023self}). 

\subsection{Experimental Results}
Table~\ref{tab:results} presents comprehensive evaluation results across different domains and model architectures. Several key observations emerge from our experiments: \ourmethod demonstrates consistent improvements across specialized domains, with Llama-3.1-8b-it showing notable gains in Finance (+2.7\%), Music (+2.8\%), and Movie (+3.3\%) domains, and particularly strong performance in medical domains (BioASQ +19.4\%, PubMedQA +10.8\%). All three model architectures benefit from our approach: Llama-3.1-8b-it achieves the highest overall accuracy (31.9\%), Llama-2-7b-chat shows remarkable adaptation capability in medical domains (+17.7\% on BioASQ), and ChatGLM-3-6b demonstrates significant improvements in PubMedQA (+25.0\%) and consistent gains across CRAG domains (+3.7\% overall). While both CoT and ICL show some improvements over the naive-rag models, \ourmethod consistently outperforms these baselines in specialized domains, with the only exception being Open domain tasks where CoT occasionally shows stronger performance, particularly with Llama-3.1-8b-it (45.8\% vs 41.5\%).

\begin{table*}[t]
    \centering
    \caption{Comparison of our method with state-of-the-art pretrained RAG models. All models utilize Llama-2-7b-chat as the backbone.}
    \label{tab:results_llama2}
    \begin{tabular}{l@{\hspace{8pt}}cccccccc}
    \toprule
    & \multicolumn{6}{c}{\textbf{CRAG}} & \multicolumn{2}{c}{\textbf{Medical}} \\
    \cmidrule(lr){2-7} \cmidrule(l){8-9}
    \textbf{Model} & \textbf{Finance} & \textbf{Sports} & \textbf{Music} & \textbf{Movie} & \textbf{Open} & \textbf{Overall} & \textbf{BioASQ} & \textbf{PubMedQA} \\
    \midrule
    Naive-RAG (Llama-2-7b-chat) & 14.7 & 23.2 & 36.5 & 30.4 & 39.2 & 27.8 & 54.1 & 47.6 \\
    \midrule
    Ret-Robust & 14.6 & 20.6 & 33.2 & 32.4 & 33.5 & 26.1 & 24.7 & 28.4 \\
    RAAT & 13.4 & 18.1 & 28.6 & 25.2 & 31.7 & 22.7 & 64.9 & 46.6 \\
    Self-rag & 11.4 & 19.8 & 22.5 & 20.9 & 26.7 & 19.8 & 57.1 & 43.4 \\
    \midrule
    \rowcolor{gray!20}\ourmethod & \textbf{16.4} & \textbf{25.8} & \textbf{40.7} & \textbf{33.8} & \textbf{41.1} & \textbf{30.5} & \textbf{71.8} & \textbf{54.0} \\
    \bottomrule
    \end{tabular}
    \end{table*}

Table~\ref{tab:results_llama2} presents a performance comparison between different RAG models across various domains. The compared pretrained RAG models (Ret-Robust, RAAT, and Self-RAG) are general-domain models and were not further fine-tuned on the specific downstream task domains used in our evaluation. Our method, \ourmethod, by contrast, adapts at test-time using unlabeled passages from the target domain. Notably, three of the models (Ret-robust, RAAT, and Self-rag) are pre-trained RAG models based on Llama-2. Despite Ret-robust using the larger Llama-2-13b as its base, and RAAT and Self-rag using Llama-2-7b, all three pre-trained RAG models perform worse than the Llama-2-7b-chat model (which achieves 27.8\% overall accuracy). This underperformance is consistent across most domains, with only RAAT showing strength in the BioASQ medical domain (64.9\%). The results suggest that current RAG pre-training methods have limited generalization capabilities, as they fail to match or exceed the performance of the base model, even when using larger model architectures. \ourmethod outperforms all other models across all domains, demonstrating the effectiveness of its approach compared to existing RAG pre-training methods.


\noindent\textbf{The effectiveness of segment-based adaptation} We compare our segment-based approach (splitting passages into prefix-suffix pairs) with a baseline that does not segment the passage, where we perform next-token prediction on the entire passage without segmentation. The results in Table~\ref{tab:segmentation} demonstrate that the segmentation strategy yields consistent performance gains across all model architectures: +1.1\% for Llama-3.1-8b-it, +0.4\% for Llama-2-7b-chat, and +0.7\% for ChatGLM-3-6b. We attribute these improvements to the front-to-back prediction task better aligning with natural language understanding compared to token-by-token prediction, enabling more effective parameter updates. The larger improvement observed with Llama-3.1-8b-it (+1.1\%) suggests that higher-capacity models may particularly benefit from structured adaptation approaches.

\begin{table}[h]
\centering
\caption{The effectiveness of  segmentation.}
\begin{tabular}{l@{\hspace{3pt}}c@{\hspace{3pt}}c@{\hspace{3pt}}c}
\toprule
\textbf{\scriptsize Strategy} & \textbf{\scriptsize Llama-3.1-8b-it} & \textbf{\scriptsize Llama-2-7b-chat} & \textbf{\scriptsize ChatGLM-3-6b} \\
\midrule
\ourmethod & 31.9 & 30.5 & 25.7 \\
wo seg & 30.8 & 30.1 & 25.0 \\
\bottomrule
\end{tabular}
\label{tab:segmentation}
\end{table}

\begin{figure}[h]
  \centering
  \includegraphics[width=1.0\linewidth]{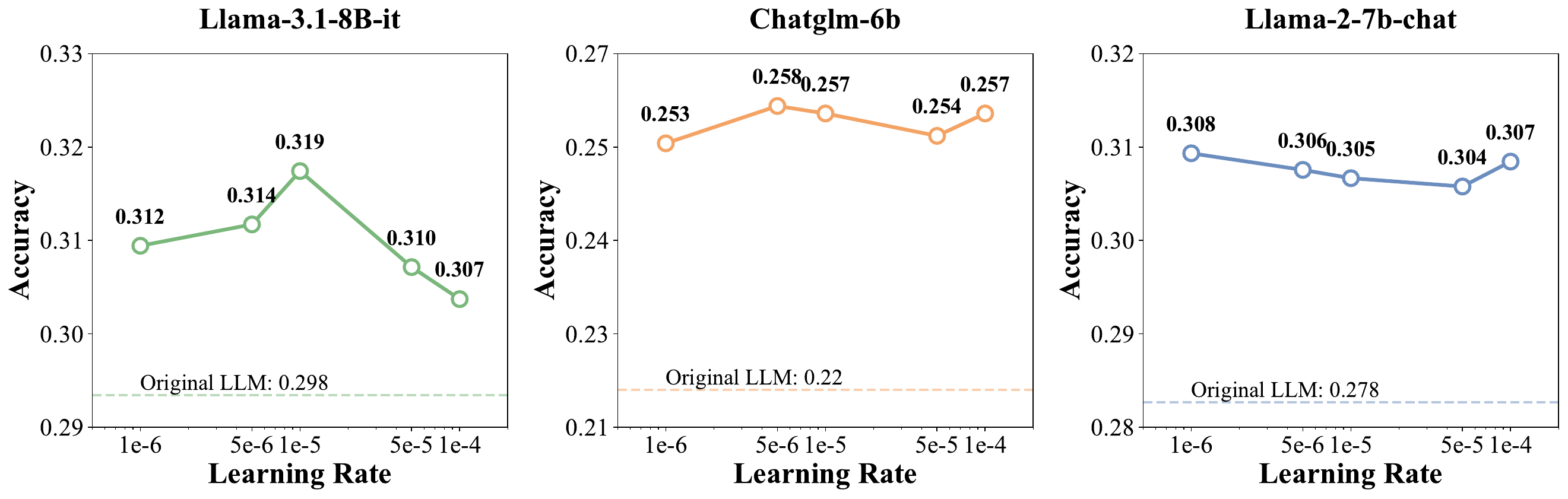}
  \caption{Accuracy vs. Learning Rate}
  \label{fig:accuracy_vs_lr}
\end{figure}

\begin{figure}[h]
  \centering
  \includegraphics[width=1.0\linewidth]{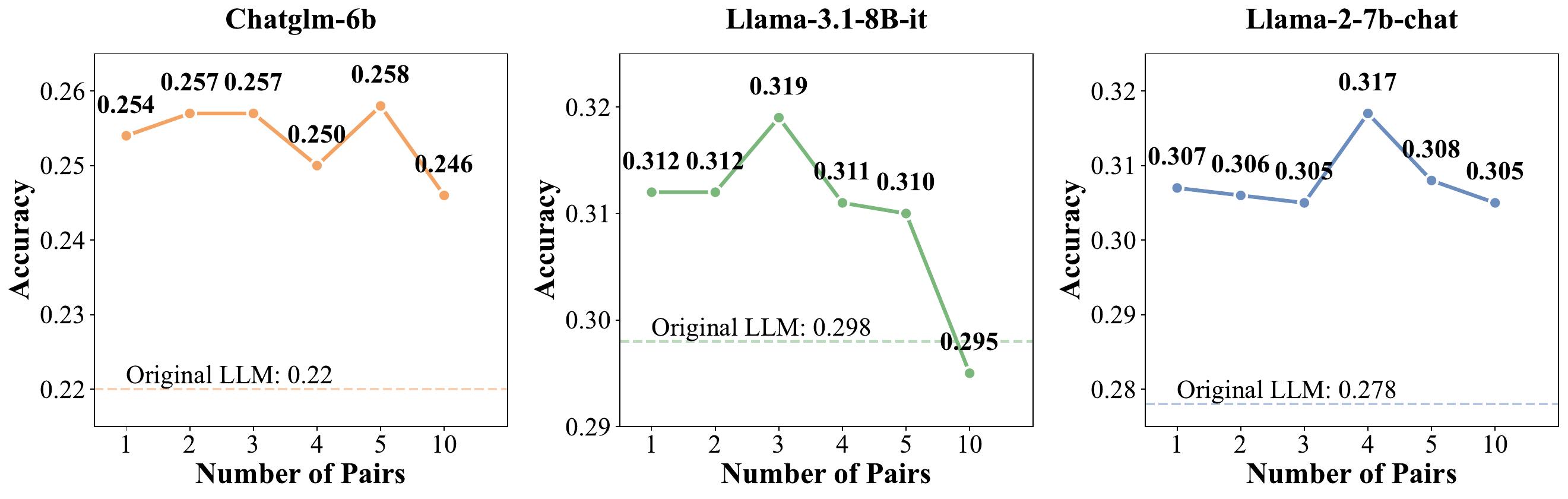}
  \caption{Accuracy vs. Number of Adaptation Pairs}
  \label{fig:accuracy_vs_pairs}
\end{figure}

\noindent\textbf{Hyperparameter Analysis.}
We also conduct hyper-parameter analysis about the number of adaptation pairs and learning rate. 
As shown in Figure~\ref{fig:accuracy_vs_lr} and Figure~\ref{fig:accuracy_vs_pairs}, our experiments show that learning rates between 1e-6 to 1e-5 provide optimal performance, with 3-5 adaptation pairs yielding the best balance between effectiveness and efficiency. However, the performance of \ourmethod is not sensitive to the number of adaptation pairs and learning rate. In almost all cases, the performance of \ourmethod is better than the naive-rag model.

\noindent\textbf{On the computation efficiency.}
To evaluate the computational overhead of our approach, we measure the total inference time across different configurations and compare it with baseline methods. The computation time reported in Table~\ref{tab:computation} was measured on the CRAG dataset, processing 2,706 queries on one NVIDIA A100 GPU. For each query, up to 5 passages were used for \ourmethod adaptation. Table~\ref{tab:computation} shows the total and average inference times for different numbers of adaptation pairs (1-5), compared against Chain-of-Thought (CoT) and the original model without adaptation. The results are based on processing 2,706 queries from the CRAG dataset.

\begin{table}[h]
\centering
\footnotesize
\caption{Computation time analysis. Total and average inference times (in seconds) for different numbers of adaptation pairs, compared against CoT and vanilla model.}
\label{tab:computation}
\begin{tabular}{@{}l@{\hspace{4pt}}c@{\hspace{4pt}}c@{\hspace{4pt}}c@{\hspace{4pt}}c@{\hspace{4pt}}c@{\hspace{4pt}}c@{\hspace{4pt}}c@{}}
\toprule
\textbf{Metric} & \textbf{1pair} & \textbf{2pair} & \textbf{3pair} & \textbf{4pair} & \textbf{5pair} & \textbf{CoT} & \textbf{Naive-RAG} \\
\midrule
 Total (s) & 4,740 & 5,723 & 6,621 & 7,001 & 7,023 & 11,688 & 961 \\
Avg (s) & 1.75 & 2.11 & 2.45 & 2.59 & 2.60 & 4.32 & 0.36 \\
\bottomrule
\end{tabular}
\end{table}

While our method does introduce additional computational overhead compared to the original model, it remains significantly more efficient than CoT. The average processing time per query ranges from 1.75s (1-pair) to 2.60s (5-pair), which is substantially lower than CoT's 4.32s. This demonstrates that \ourmethod achieves its performance improvements with reasonable computational cost, making it practical for real-world applications.

\section{Conclusion}

In this paper, we present \ourmethod, a test-time adaptation approach for retrieval-augmented generation that enables dynamic model optimization during inference. Our method introduces a simple yet effective self-supervised learning objective where the model learns to predict retrieved content, allowing automatic parameter adjustment to target domains without requiring labeled data. Through extensive experiments across six specialized domain, we demonstrate that \ourmethod achieves consistent improvements over the naive-rag system. These findings suggest that test-time adaptation is a promising direction for improving RAG systems' performance in specialized domains.

\footnotesize
\bibliographystyle{IEEEbib}
\bibliography{custom}

\end{document}